\documentclass[journal]{IEEEtran}
\usepackage{cite}
\usepackage{amsmath,amssymb,amsfonts}
\usepackage{graphicx}
\usepackage{booktabs}
\usepackage{multirow}
\usepackage{xcolor}
\usepackage{url}
\usepackage[hidelinks]{hyperref}
\graphicspath{{figures/}}

\def\BibTeX{{\rm B\kern-.05em{\sc i\kern-.025em b}\kern-.08em T\kern-.1667em\lower.7ex\hbox{E}\kern-.125emX}}

\begin{document}

\title{LURE: a Phase-Labelled Synthetic Corpus of Online-Grooming Exchanges}

\author{Emanuele~Florindi,~\IEEEmembership{Member,~IEEE,}
        Valentina~Franzoni,~\IEEEmembership{Senior~Member,~IEEE,}
        Ludovico~Guercio,
        Alfredo~Milani,~\IEEEmembership{Senior~Member,~IEEE,}
        and~Mattia~Polticchia,~\IEEEmembership{Member,~IEEE}%
\thanks{Preprint. This work received no project funding; it was supported by the EMORE Lab, University of Perugia, Italy.}%
\thanks{V. Franzoni and L. Guercio are with the EMORE Lab, Department of Mathematics and Computer Science, University of Perugia, 06123 Perugia, Italy. V. Franzoni is also with the Department of Computer Science, Hong Kong Baptist University, Kowloon Tong, Hong Kong (corresponding author, e-mail: valentina.franzoni@unipg.it).}%
\thanks{E. Florindi and M. Polticchia are with the LESDIT PhD School, Learning Science and Digital Technologies, University of Modena and Reggio Emilia, Reggio Emilia, Italy.}%
\thanks{A. Milani is with the Department of Human Sciences, Link Campus University, Rome, Italy.}%
\thanks{ORCID: E. Florindi 0009-0002-5420-0326; V. Franzoni 0000-0002-2972-7188; A. Milani 0000-0003-4534-1805; M. Polticchia 0009-0005-6175-9832.}}

\markboth{Preprint}%
{Florindi \MakeLowercase{\textit{et al.}}: LURE: a Phase-Labelled Synthetic Corpus of Online-Grooming Exchanges}

\maketitle

\begin{abstract}
Phase-annotated conversational data is essential for studying the progression of online grooming of minors and for developing early warning tools. However, the only publicly available source (i.e., Perverted-Justice archive, which contains a subset coded using ChatCoder) is small, outdated, and frozen. We present a synthetic corpus of groomer–victim exchanges, generated using a retrieval-augmented pipeline running on institutional hardware. This pipeline uses an openly available, eight-billion-parameter, refusal-ablated language model. The corpus was created using the five stages of the responsible-generation protocol for synthetic datasets: ontology-driven design based on the ChatCoder luring-communication scheme (eight communicative tags mapped onto four grooming phases); multi-model prompt testing; human-in-the-loop supervision; semantic and feature-level validation; and modality-aware documentation. The released corpus contains 2,121 unique exchanges across seven tags and four phases, each accompanied by a Plutchik emotion field. Validation quantifies the integrity and lexical profile of the corpus, as well as the reproduction of real seed lines, in-domain label recoverability (tag accuracy: 0.85; phase accuracy: 0.93) and its transfer to real ChatCoder offender lines, which were coded automatically at the eight-tag level (31,640 lines) and by human coders at the three-construct level (10,876 posts). Accuracy was measured for lexical and sentence-embedding features, while a real-versus-synthetic discriminator was able to separate the two domains with 91\% accuracy. The scope of the corpus includes controlled, phase-conditioned experimentation, pipeline development and augmentation alongside real data and awareness material. The corpus, its datasheet and the audit code are released with vetted access.
\end{abstract}


\section{Introduction}
\IEEEPARstart{O}{nline} grooming is the gradual process by which an adult establishes trust and emotional dependence with a child in order to facilitate sexual contact or obtain material~\cite{whittle2013,wolak2004}. This phenomenon has grown along with social networks, messaging apps, and online gaming. The average age of victims is dropping, with pre-adolescents being the most exposed group \cite{polposta2024,iwf,ncmec,11003860}. Its handling is being debated at the European level, where institutional client-side scanning has been assessed as risky \cite{tyad020} and is being reconsidered \cite{neweulaw26}; scanning by law enforcement based on well-founded suspicion or via a tool chosen and managed by the user remains the more proportionate approach. Both require models that are trained using phase-annotated conversational data.
Grooming progresses through identifiable stages of communication (contact, building trust, isolation from protective adults, desensitisation to sexual content and approach), which can be studied and recognised in chat text~\cite{olson2007luring,chatcoder}. The computational work on grooming relies on a single public source: the Perverted-Justice (PJ) archive of conversations between convicted offenders and adult volunteer decoys. This archive was used in the PAN 2012 shared task and in most subsequent detection studies \cite{pan12,groomingsurvey,vogt2021early}. The archive is in English, and the only portion with message-level communicative annotation is its ChatCoder subset (288 transcripts coded automatically with a 475-phrase dictionary and 55 coded by trained human coders) \cite{chatcoder}. Data protection laws mean that most research groups cannot access new real data on sexual communication with children \cite{gdpr}.

\begin{figure*}[!t]
\centering
\includegraphics[width=.95\textwidth]{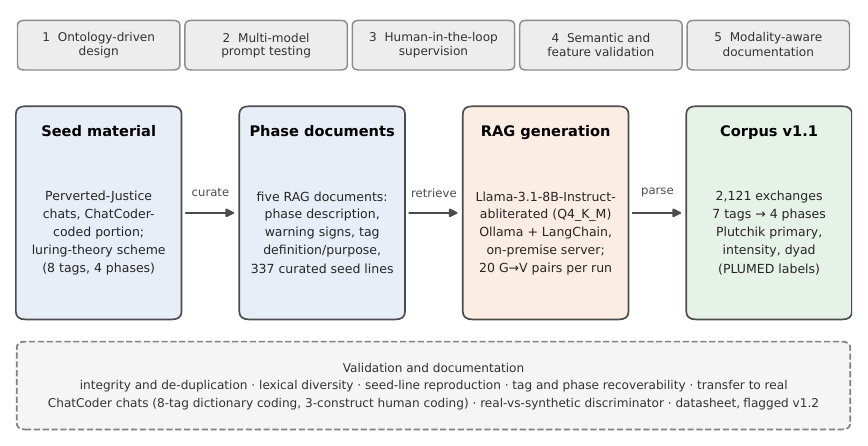}
\caption{The generation pipeline and its role in the responsible generation protocol \cite{franzoni2025responsible} (top strip). Perverted-Justice material coded by ChatCoder \cite{chatcoder} is curated into five-phase documents that anchor a locally hosted RAG chain. Example-driven, per-phase, per-tag generation emits labelled groomer–victim pairs that are parsed into the corpus and annotated with the Plutchik emotion field. The lower strip lists the validation and documentation steps of Section~\ref{sec:validation}.}
\label{fig:pipeline}
\end{figure*}

The use of Large Language Models (LLMs) to generate labelled text is an established response to data scarcity in sensitive domains \cite{dataaug,Yoo20212225}, and grooming is a natural target: the label of interest (i.e., the grooming phase) is imposed at generation time and no minors are involved. The domain also encompasses the questions that responsible generation must answer. Aligned models refuse the task, while refusal-ablated open models comply but offer no domain knowledge. The most informative phases are the most explicit, and every artefact is dual-use. The protocol of \cite{franzoni2025responsible} organises the generation of synthetic datasets under these exact conditions into five stages: ontology-driven design; multi-model prompt testing; iterative human-in-the-loop supervision; semantic and feature-level validation; and modality-aware documentation. This process is illustrated using a Plutchik-based emotion corpus and has since been applied to the PLUMED corpus \cite{amici2026plumed}. This paper uses the same protocol for online grooming and provides details of each stage, along with the measurements needed to make the resulting resource usable by others. Figure~\ref{fig:pipeline} summarises the pipeline and the five stages.

Our contributions are:
\begin{itemize}
\item This is a retrieval-augmented generation (RAG) pipeline for phase- and tag-labelled grooming exchanges. It runs on institutional infrastructure and uses openly available, refusal-ablated checkpoints, which are served by Ollama and orchestrated with LangChain. The pipeline is anchored to curated, de-identified seed documents, which are built from the ChatCoder-coded portion of PJ (Sections~\ref{sec:data}--\ref{sec:method}).
\item A corpus of 2,121 unique groomer-victim exchanges (LURE v1.1, \textit{Luring-theory Utterances for Research on grooming Exchanges}) has been released. Each exchange carries a ChatCoder communicative tag, the corresponding grooming phase, and a Plutchik emotion field. A datasheet is also available \cite{gebru2021datasheets} and flagged v1.2 (Section~\ref{sec:corpus}).
\item A validation suite covering integrity, lexical profile, seed-line reproduction and in-domain label recoverability. External validation is also provided against real ChatCoder data at two annotation levels: automatic eight-tag coding (silver) and human three-construct coding (gold). Cross-domain transfer is covered by lexical and sentence-embedding features, as well as a real versus synthetic discriminator. (Section~\ref{sec:validation}).
\end{itemize}
These measurements will provide a documented scope of use and protocol refinements for future synthetic corpora in sensitive domains. They will also inform the governance framework for producing and sharing such data. (Sections~\ref{sec:discussion}-\ref{sec:ethics}).
\section{Related Work}
Research on the detection of online grooming is shaped by the tension between the need for early-warning models and the limited availability of up-to-date, ethically shareable conversational data. While cybersecurity work has focused on preserving access and channels (e.g., age and ageing in biometrics \cite{8509614}, encrypted-traffic analysis \cite{10964328}, intrusion detection \cite{10948513}), while tools for minor protection recognise abusive content once abuse has begun \cite{9541357,11021376}; grooming mainly takes place on social networks and in online games, where remote interaction enables the crime~\cite{11003860}.

\subsection{Grooming and abuse detection in text}
Automated grooming detection has largely been approached as a form of supervised text classification using PJ-derived corpora, incorporating lexical, stylometric and, more recently, Transformer features \cite{chatcoder,pj,pan12,groomingsurvey}. Communication scholars proposed stage models of luring \cite{olson2007luring}; the ChatCoder project categorised them into eight utterance-level categories and created a coding dictionary. This was applied automatically to 288 PJ transcripts and, for a smaller set, by trained human coders at the level of the theory's constructs~\cite{chatcoder}. More recent work has shifted towards early detection settings and privacy-preserving training \cite{vogt2021early,chehbouni2024federated}, and has begun to explore the use of LLMs as estimators of grooming risk \cite{prosser2024helpful}. Cyberbullying and the detection of predatory behaviour in gaming chats share the same data constraints  \cite{Cheong2015220,Murnion2018197}, and grooming through gaming platforms follows recognisable, platform-specific patterns \cite{ran2021,florindi2024}. Discourse analysis describes the language used for grooming in computer-mediated settings \cite{lorenzodus2016understanding,winters2017sexual}. The corpus presented here adds a phase- and tag-labelled synthetic resource documented against the ChatCoder material.

Three studies delimit the space in which LURE sits, and a direct comparison clarifies what it contributes. The survey of Borj \emph{et al.}\ \cite{groomingsurvey} maps the field from the data side: it reviews the corpora on which our grooming detection study is built and the techniques applied to them, and documents that the available resources carry a conversation- or author-level decision (i.e., predator or not) over PJ, so that the communicative progression inside a conversation remains outside the label space. LURE is positioned at that point: it supplies the message-level layer the survey reports as scarce, with a tag and a phase on every exchange, and it reports where its distributions sit with respect to the real ChatCoder coding. Cheong \emph{et al.}\ \cite{Cheong2015220} work with real chats from a children's social game, which gives their study the platform realism and the age profile that PJ lacks; those logs were obtained under an agreement with the operator and stay with the research team, which is the condition that makes such corpora difficult to reuse. Besides not being labeled for grooming content, such data should be a good starting point for children chats on games. LURE addresses the same reuse problem from the other side: the exchanges are generated rather than collected, so the resource can be shared under a data-use agreement, and the external validation in Section~\ref{sec:external} states how far the generated text stands from the real material. Murnion \emph{et al.}\ \cite{Murnion2018197} assemble 126,091 in-game messages from 14,930 match replays and score them for cyberbullying, demonstrating the scale that automatic harvesting of a public game channel reaches on a harm that is observable in open chat; grooming is pursued in private channels and is identified only after a judicial process, which is the reason its corpora are counted in hundreds of conversations. LURE offers a complementary route to volume for this second setting, with generation under human supervision and a release documented for controlled benchmarking, focusing on grooming instead of cyberbulling.

\subsection{Responsible generation of synthetic corpora}
Data generation driven by LLM with retrieval augmentation is a standard tool in low-resource and privacy-sensitive domains \cite{rag,Yoo20212225,dataaug}. The properties on which a synthetic corpus must be measured are identified by surveys. These properties include diversity relative to the generator's prompt, distributional fidelity to the target phenomenon, and transfer of what is learned on synthetic text to real text \cite{li2023synthetic,shumailov2024collapse}. Here, retrieval is used to guide creative generation towards real examples, adding the requirement to measure the reproduction of retrieved material. Data statements and datasheets provide the documentation framework \cite{bender2018data,gebru2021datasheets}; the followed protocol \cite{franzoni2025responsible} organises these requirements into five stages, which are reflected in the structure of Sections~\ref{sec:data}-\ref{sec:validation}.

\subsection{Emotion annotation}
Grooming is an emotional process, and affective computing provides text-based methods for recognising emotions \cite{affective1,affective2}, built on categorical models such as Ekman's basic emotions \cite{ekman1992} or Plutchik's wheel \cite{plutchik2001}, and on Plutchik-aligned resources such as the NRC lexicon \cite{nrclexicon}. The corpus contains a Plutchik emotional field for each exchange, which was produced using a transformer classifier that was transferred and learnt using a Plutchik-based corpus \cite{amici2026plumed}. The emotional field is part of the released schema and will be analysed in other studies.

\section{Stage 1: Ontology-Driven Design}
\label{sec:data}
\subsection{Source material}
The PJ archive contains 622 one-to-one conversations between offenders and adult decoys posing as minors on platforms such as Yahoo and MeetMe. The community copy used here has had the usernames, timestamps, and moderator comments removed. These conversations took place between 2004 and 2016 \cite{pj}. The ChatCoder project distilled a luring-communication scheme with three theoretical constructs from this archive: deceptive trust development, grooming, (communicative desensitisation and reframing), and isolation/approach. Eight utterance-level tags were also identified (see Table~\ref{tab:scheme}) \cite{olson2007luring,chatcoder}. Two coded resources distributed with the project serve as real references in Section~\ref{sec:external}: 288 transcripts that were coded automatically by the ChatCoder dictionary at the eight-tag level (177,366 offender lines, of which 35,392 contain a dictionary phrase and 31,640 carry exactly one tag) and 55 transcripts coded by trained human coders at the construct level (11,204 offender posts with a construct code, of which the 10,876 with at least two word tokens are used in the experiments). The dictionary coding is keyword-based. For instance, an offender username containing the substring 'bra' triggered the desensitisation tag on every line of that transcript. The dictionary coding is treated as silver, while the human coding is treated as gold. Each is used at its own granularity.

\begin{table*}[!t]
\caption{Annotation scheme: ChatCoder communicative tags, luring-theory construct, generation phase context and released phase label.}
\label{tab:scheme}
\centering\footnotesize
\begin{tabular*}{\textwidth}{@{\extracolsep{\fill}}llll@{}}
\toprule
\textbf{Tag} & \textbf{Construct} & \textbf{Generation context} & \textbf{Released phase} \\
\midrule
Activities & deceptive trust & Targeting \& Gaining Trust & Targeting \& Gaining Trust \\
Personal Information & deceptive trust & Targeting \& Gaining Trust & Targeting \& Gaining Trust \\
Relationship & deceptive trust & Targeting; Fulfilling Needs & (merged, see Sec.~\ref{sec:corpus}) \\
Compliment & deceptive trust & Fulfilling Needs & Fulfilling Needs \\
Reframing & grooming & Isolation & Isolation \\
Communicative Desensitization & grooming & Isolation & Isolation \\
Isolation & isolation/approach & Isolation; Control & Isolation \\
Approach & isolation/approach & Sexualizing; Control & Sexualizing the Relationship \\
\bottomrule
\end{tabular*}
\end{table*}

\subsection{Phase scheme and tag-to-phase mapping}
The design stage of the protocol establishes the label ontology prior to the creation of any prompts. Stage models of sexual abuse and online luring \cite{finkelhor1984,ward2001,whittle2013} describe a progression involving contact, friendship, risk assessment, special bonding, trust-testing missions, secrecy and luring, which exploits classic persuasion principles \cite{cialdini2009}. Generation is organised in five phase contexts: \emph{Targeting and Gaining Trust}, \emph{Fulfilling Needs}, \emph{Isolation}, \emph{Sexualizing the Relationship and Abuse} and \emph{Control}. The released label set consists of four primary phases, with every tag mapping deterministically to one of these phases (see Table~\ref{tab:scheme}). The control context, whose tags all belong to the other phases, is folded into them. The isolation phase precedes explicit sexual content and is defined by emotional rather than sexual material. This makes it the operational focus of early-warning research.

\subsection{Emotion field}
Each exchange carries a Plutchik emotion label (i.e., one of the eight primary emotions or \textit{neutral}) assigned in a separate annotation process using the \texttt{all-Roberta-large-v1 model}, which was trained on a Plutchik-grounded textual corpus \cite{amici2026plumed,plutchickTextCorpus}. This field is released as a silver-standard annotation and is documented as such in the datasheet.

\section{Stage 2: Multi-Model Prompt Testing}
\label{sec:method}
\subsection{Infrastructure and models}
Generation ran on the \textit{UniNuvola} computing service of the University of Perugia, on a node with an AMD EPYC 9334 CPU (2 cores allocated), 32 GB of RAM and an NVIDIA A30 GPU (24 GB), with models served by Ollama \cite{Ollama2024} and orchestrated with LangChain \cite{Chase2022LangChain} in Python; prompts and outputs stayed on institutional infrastructure. Three openly distributed refusal-ablated checkpoints were tested: \textit{Llama-3.2-3B-Instruct-abliterated} \cite{Huihui2025Llama32Abliterated}, \textit{Meta-Llama-3.1-8B-Instruct-abliterated} (Q4\_K\_M) \cite{Labonne2024Llama31Abliterated} and \textit{DeepSeek-R1-Distill-Llama-8B-Abliterated} (IQ4\_XS) \cite{Huihui2025DeepSeekAbliterated}. Abliteration removes the refusal direction from the residual stream \cite{arditi2024refusal}, so that these checkpoints comply with the task; domain knowledge about grooming is supplied through the prompt and the retrieval base.

\subsection{Knowledge base}
The RAG knowledge base consists of five phase documents (\textit{targeting\_and\_gaining\_trust}, \textit{fulfilling\_needs}, \textit{isolation}, \textit{sexualize\_the\_relationship\_and\_abuse}, \textit{control}) with a fixed structure: a natural-language description of the phase and its real-world warning signs; per-tag sections with the tag definition and its manipulative purpose; and de-identified offender lines selected manually from the ChatCoder-coded PJ portion (337 seed lines of at least three tokens in total). Retrieval for each call is restricted to the document matching the target (phase, tag) pair, following the standard embed--index--retrieve flow.

\subsection{Prompt architecture}
Each call concatenates three components. The \textit{system prompt} provides in-context domain grounding in the form of the two-stage luring model and its six persuasion principles, the seven-stage online sequence, gaming-specific patterns and the phase/tag scheme with its associated output conventions. The \textit{RAG prompt} instructs the model to use the retrieved examples as inspiration for creating semantically similar messages that differ in their surface form. The \textit{question prompt} specifies the task. In dialogue mode, it also receives the accumulated conversation, enabling each phase to build on the previous one. Components that address residual refusal (e.g., the purpose statement and role framing), are described at this level and kept out of the release (see Section~\ref{sec:ethics}).

\subsection{Generation strategies tested}
\begin{enumerate}
\item \textbf{Direct dialogue.} A single prompt requests a complete grooming conversation, divided into phases. The outputs were brief and only partially covered the grooming pattern; the 3B model declined the task in most runs (e.g.\ \emph{``I can't provide information or guidance on illegal or harmful activities, especially those that could endanger a child''}).
\item \textbf{Phase-by-phase dialogue.} The conversation is generated one phase at a time, with feedback on previous phases. Phase adherence and coherence improved, but residual behaviours included abrupt topic shifts, the recurrence of covered topics, limited variation in the victim's voice and sparse output during the explicit phases. The RAG-optimised checkpoints (Llama-3.2 and DeepSeek-R1) tended to produce retrieved seed lines with few modifications in one archived run, six of the 38 generated messages were seed lines. \textit{Llama-3.1-8B-Instruct-abliterated} produced the most stable output and was retained.
\item \textbf{Example-driven per-tag pairs (adopted).} The model produces 20 groomer-to-victim message pairs in machine-readable format \texttt{message|tag|phase} per run for a given tag within a given phase. Retrieval is restricted to the matching document and the corpus is labelled by construction. The task portion of the question prompt reads:
\begin{quote}\footnotesize\ttfamily
Generate 20 samples related to Tag \{tag\} for a chat between groomer and victim in an online platform. For each sample, generate a message from the groomer, and the reply from victim in the following format:\\
groomer message|tag|phase\\
victim message|tag|phase\\
Retrieve context only from the document that corresponds to phase "\{phase\}" and tag "\{tag\}". [\ldots]
\end{quote}
\end{enumerate}
Table~\ref{tab:strategies} summarises the three strategies. Eight per-tag output files were produced and parsed into exchanges, each consisting of one groomer message and its victim reply.

\begin{table}[!t]
\caption{Comparison of the three generation strategies across the evaluated locally hosted refusal-ablated models.}
\label{tab:strategies}
\centering\footnotesize
\begin{tabular}{@{}p{1.8cm}p{2.7cm}p{3.2cm}@{}}
\toprule
\textbf{Strategy} & \textbf{Observed quality} & \textbf{Residual behaviours} \\
\midrule
S1: direct dialogue & Short, partial grooming patterns & Refusal (3B model); no phase structure \\
S2: phase-by-phase dialogue & Coherent, phase-adherent (RAG anchoring) & Topic shifts; repetition; flat victim voice; sparse output on explicit phases \\
S3: example-driven per-tag pairs (\textbf{adopted}) & Phase-consistent, labelled by construction & Limited lexical variation; seed-line reproduction in RAG-optimised models \\
\bottomrule
\end{tabular}
\end{table}

\section{Stage 3: Human-in-the-Loop Supervision}
\label{sec:hitl}
Human supervision was introduced at four points in the pipeline:
\begin{enumerate}
    \item \emph{Seed curation}:
    the seed lines were selected manually from the ChatCoder-coded section and organised by phase and tag.
    \item \emph{Document engineering}: early versions of the phase documents only retrieved basic example lists; adding explanatory text at phase and tag level improved the relevance of the retrieved context.
    \item \emph{Prompt refinement}: the system and question prompts evolved across test runs to fit the models' context budget while retaining the domain description.
    \item \emph{Output review}: approximately one quarter of the generated samples were inspected for message plausibility and label consistency. Disagreements were resolved by majority vote among four researchers. The strategy comparison in Table~\ref{tab:strategies} is based on this inspection. 
\end{enumerate}
The release records show where supervision was applied, as well as where sampling was used instead of an exhaustive approach.

\section{The Released LURE Corpus}
\label{sec:corpus}
\subsection{Versions}
\textbf{v1.0} is the corpus as generated and parsed: 2,579 exchanges over eight tag files. Integrity checking (Section~\ref{sec:integrity}) established that the \textit{Relationship} output file had the same content as the \textit{Compliment} file, and identified 128 further exact duplicates. \textbf{v1.1}, the released version, keeps one copy of every exchange: 2,121 unique exchanges over seven tags and four phases (Table~\ref{tab:corpus}). \textbf{v1.2} adds two Boolean columns: \texttt{seed\_verbatim} (the groomer message equals a seed line after normalisation; 151 exchanges) and \texttt{seed\_4gram\_full} (every 4-gram of the groomer message occurs in the seed set; 162 exchanges; union 167), so that users can select the subset in which no seed line is reproduced.

\begin{table}[!t]
\caption{Composition of the corpus by phase and tag: exchanges as generated (v1.0) and as released (v1.1).}
\label{tab:corpus}
\centering\footnotesize
\begin{tabular}{@{}llrr@{}}
\toprule
\textbf{Phase} & \textbf{Tag} & \textbf{v1.0} & \textbf{v1.1} \\
\midrule
Targeting/Gaining Trust & Activities & 316 & 307 \\
Targeting/Gaining Trust & Personal Info & 357 & 357 \\
Targeting/Gaining Trust & Relationship & 330 & --- \\
Fulfilling Needs & Compliment & 351 & 315 \\
Isolation & Reframing & 194 & 158 \\
Isolation & Isolation & 365 & 355 \\
Isolation & Desensitization & 358 & 356 \\
Sexualizing the Relationship & Approach & 308 & 273 \\
\midrule
\multicolumn{2}{@{}l}{Targeting / Fulfilling / Isolation / Sexualizing} & \multicolumn{2}{r}{664 / 315 / 869 / 273} \\
\textbf{Total} & & \textbf{2,579} & \textbf{2,121} \\
\bottomrule
\end{tabular}
\end{table}

\subsection{Schema and documentation}
Each row carries \texttt{exchange\_id}; \texttt{sentence} (the exchange as ``G: \ldots\ | V: \ldots''); \texttt{groomer\_text}, \texttt{victim\_text}; \texttt{tag}, \texttt{tag\_label}; \texttt{phase}, \texttt{phase\_label}; \texttt{emotion}; token counts; and \texttt{source\_id} (originating per-tag file and line). Table~\ref{tab:stats} gives the global statistics, Table~\ref{tab:emotions} the distribution of the emotion field, and Table~\ref{tab:examples} representative groomer messages curated for responsible disclosure. The datasheet \cite{gebru2021datasheets} records provenance (fully synthetic; RAG-anchored generation with manually curated seed examples), generator (\textit{Llama-3.1-8B-Instruct-abliterated}), language (English), the supervision points of Section~\ref{sec:hitl}, label status (tags and phases by construction; emotion as silver annotation), the validation figures of Section~\ref{sec:validation}, known biases (seed-selection subjectivity, generator-distribution artefacts, simplified victim voice), the withheld components and the access regime.

\begin{table}[!t]
\caption{Global statistics of the corpus as generated (v1.0) and as released (v1.1). Duplicates are counted on the exchange string after lower-casing and punctuation removal; distinct-$n$ is the ratio of unique to total $n$-grams; lengths are word tokens.}
\label{tab:stats}
\centering\footnotesize
\begin{tabular}{@{}lrr@{}}
\toprule
\textbf{Statistic} & \textbf{v1.0} & \textbf{v1.1} \\
\midrule
Exchanges (groomer--victim pairs) & 2,579 & 2,121 \\
Tags / phases & 8 / 4 & 7 / 4 \\
Exact duplicate exchanges & 458 & 0 \\
Near-duplicate exchanges (normalised) & 466 & 8 \\
Tokens (exchange level) & 40,657 & 37,039 \\
Vocabulary (word types) & 1,971 & 1,971 \\
Type--token ratio & 0.043 & 0.053 \\
Distinct-2 / distinct-3 & 0.25 / 0.45 & 0.30 / 0.56 \\
Mean length groomer / victim & 8.6 / 7.1 & 8.6 / 6.9 \\
Exchanges containing emoji (\%) & 82 & 80 \\
Groomer messages identical to a seed line (\%) & --- & 7.1 \\
Seed material: PJ conversations / ChatCoder phrases & \multicolumn{2}{c}{622 / 475} \\
Generation batch size & \multicolumn{2}{c}{20 pairs per run} \\
\bottomrule
\end{tabular}
\end{table}

\begin{table}[!t]
\caption{Distribution of the Plutchik emotion field (eight primary emotions plus neutral) in v1.0 and v1.1.}
\label{tab:emotions}
\centering\footnotesize
\begin{tabular}{@{}lrrr@{}}
\toprule
\textbf{Emotion} & \textbf{v1.0} & \textbf{v1.1} & \textbf{v1.1 (\%)} \\
\midrule
joy & 659 & 468 & 22.1 \\
trust & 546 & 405 & 19.1 \\
sadness & 441 & 404 & 19.0 \\
anticipation & 383 & 340 & 16.0 \\
surprise & 196 & 172 & 8.1 \\
neutral & 155 & 151 & 7.1 \\
fear & 110 & 101 & 4.8 \\
anger & 56 & 51 & 2.4 \\
disgust & 33 & 29 & 1.4 \\
\midrule
\textbf{Total} & \textbf{2,579} & \textbf{2,121} & \textbf{100.0} \\
\bottomrule
\end{tabular}
\end{table}

\begin{table}[!t]
\caption{Representative synthetic groomer messages with tag and phase, curated for responsible disclosure.}
\label{tab:examples}
\centering\footnotesize
\begin{tabular}{@{}p{3.6cm}p{1.9cm}p{2.1cm}@{}}
\toprule
\textbf{Generated message} & \textbf{Tag} & \textbf{Phase} \\
\midrule
``That's understandable. Do you have any favorite games?'' & Activities & Targeting and Gaining Trust \\
``Are your parents divorced?'' & Personal Information & Targeting and Gaining Trust \\
``Your skills are insane! You're the best gamer I've ever seen!'' & Compliment & Fulfilling Needs \\
``I can be your teacher, and we can learn together'' & Reframing & Isolation \\
``Don't worry about your mom's calls, I'm here for you'' & Isolation & Isolation \\
``Dont u think 18+ games r more fun?'' & Communicative Desensitization & Isolation \\
``Can we meet in a private room?'' & Approach & Sexualizing the Relationship \\
\bottomrule
\end{tabular}
\end{table}

\section{Stage 4: Validation}
\label{sec:validation}
All experiments use TF-IDF over word unigrams and bigrams with class-balanced multinomial logistic regression (lexical), and mean-pooled frozen \texttt{all-roberta-large-v1}  (1,024 dimensions) with the same classifier (semantic), where stated. In-domain figures are based on 5-fold stratified cross-validation (mean ± s.d.); real-data folds are grouped by transcript.

\subsection{Integrity, lexical profile and seed reproduction}
\label{sec:integrity}
\textbf{Integrity.} A byte-wise comparison of the eight raw output files, along with normalised matching of exchanges, identified the Relationship/Compliment duplication (330 exchanges), as well as 128 further exact duplicates. v1.1 retains one copy of each. Within v1.1, 209 groomer messages and 313 victim replies occur in more than one exchange with a different partner's response. Both are retained because the unit of the corpus is the exchange.

\begin{figure*}[!t]
\centering
\includegraphics[width=.95\textwidth]{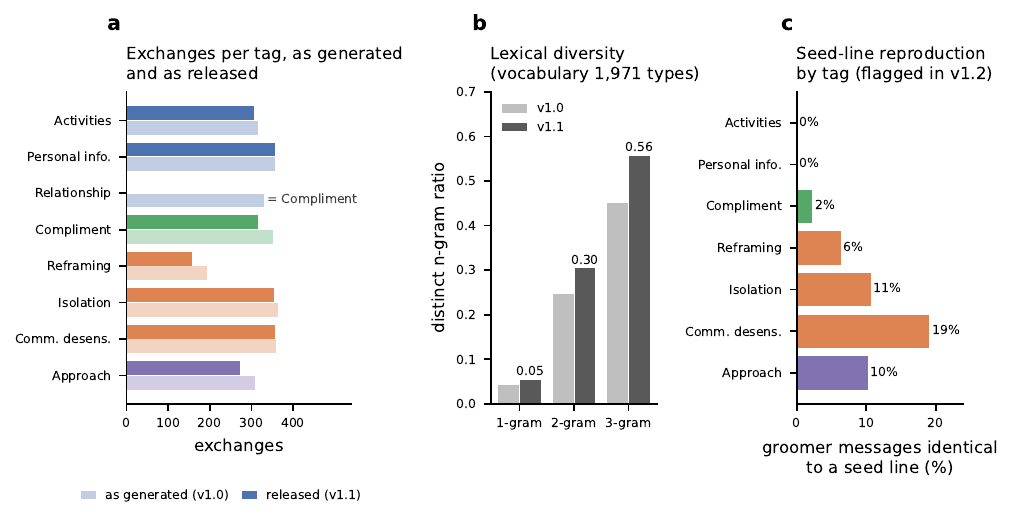}
\caption{Integrity, lexical profile and seed reproduction. (a) Exchanges per tag as generated (v1.0) and as released (v1.1); colours follow the phase (blue Targeting, green Fulfilling Needs, orange Isolation, purple Sexualizing). (b) Distinct-$n$ ratios of the two versions. (c) Share of groomer messages identical to a seed line, by tag; these exchanges carry the \texttt{seed\_verbatim} flag in v1.2.}
\label{fig:audit}
\end{figure*}

\textbf{Lexical profile.} The released corpus contains 1,971 word types across 37,039 tokens, giving a type-to-token ratio of 0.053. The distinct 1/2/3 ratios are 0.05, 0.30 and 0.56 (see Fig.~\ref{fig:audit}b). On average, groomer messages contain 8.6 tokens, while victim replies contain 6.9. Emojis are present in at least one of the messages in 80\% of exchanges. The ten most frequent groomer openings ('Have you ever', 'Do you have', 'Do you like', 'What's your', 'You have such', etc.) account for 470 out of 2,121 messages, which is a characteristic of the generator's register.

\textbf{Seed-line reproduction.} After normalisation, 151 groomer messages (7.1\%) equal a seed line against the 337 seed lines. Of these, 162 (7.6\%) consist entirely of seed 4-grams and 18.5\% share at least one 4-gram with a seed line (mean 4-gram containment 0.11). By tag (Fig.~\ref{fig:audit}c), 19\% of Communicative Desensitization, 11\% of Isolation and 10\% of Approach messages equal a seed line, 6\% of Reframing, 2\% of Compliment, and none of Activities or Personal Information. Victim replies contain no seed line. Against the ChatCoder dictionary reconstructed from the coded transcripts (299 of the 475 phrases were recoverable), 53\% of synthetic groomer messages contain a dictionary phrase, compared with 20\% of real offender lines; where a phrase is present, its ChatCoder category coincides with the assigned tag in 35\% of cases. All exchanges reproducing a seed line are flagged in v1.2.

\subsection{In-domain label recoverability}
\begin{figure*}[!t]
\centering
\includegraphics[width=.95\textwidth]{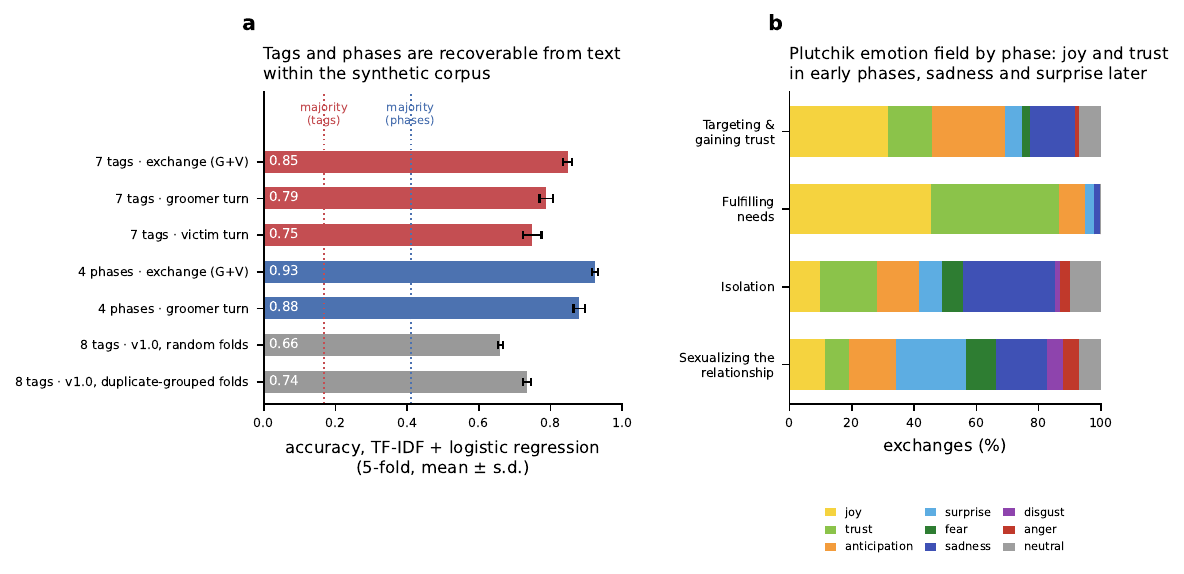}
\caption{In-domain validation. (a) Tag and phase recoverability from text with the lexical model (5-fold, mean $\pm$ s.d.); grey rows are the v1.0 corpus before de-duplication. (b) Distribution of the Plutchik emotion field by phase ($n$ = 2,121 exchanges).}
\label{fig:recov}
\end{figure*}

The label recoverability measure indicates how consistently tags and phases are expressed in the corpus text (Table~\ref{tab:recov}, Fig.~\ref{fig:recov}). The lexical model achieves an accuracy of 0.848 ± 0.013 for the seven tags and 0.925 ± 0.009 for the four phases when the entire exchange is considered. Groomer turns alone achieve an accuracy of 0.788/0.880, while victim turns alone achieve an accuracy of 0.749. A fine-tuned \texttt{bert-base-uncased} (max length 64, batch 16, AdamW $2\times10^{-5}$, 3 epochs) on the same phase folds reaches 0.918  ±  0.014, with Isolation recall 0.890 and precision 0.928; the semantic model reaches 0.829  ±  0.007 on phases and 0.768  ±  0.019 on tags. Here, lexical and transformer models are equivalent, which is consistent with the idea that per-tag generation produces text that is characteristic of each label in terms of its lexical content. The majority-class references are 0.168 (tags) and 0.410 (phases). The v1.0 rows document the effect of the duplicated file: the eight-tag accuracy is 0.660 with random folds and 0.735 with duplicate-grouped folds; the two identical tags remain merged in both cases.

\begin{table}[!t]
\caption{In-domain recoverability of tags and phases from text (stratified 5-fold, mean $\pm$ s.d.).}
\label{tab:recov}
\centering\footnotesize
\begin{tabular}{@{}lcc@{}}
\toprule
\textbf{Setting} & \textbf{Accuracy} & \textbf{Macro-F1} \\
\midrule
7 tags, exchange (G+V), lexical & 0.848 $\pm$ 0.013 & 0.844 $\pm$ 0.013 \\
7 tags, groomer turn only, lexical & 0.788 $\pm$ 0.019 & 0.783 $\pm$ 0.021 \\
7 tags, victim turn only, lexical & 0.749 $\pm$ 0.026 & 0.745 $\pm$ 0.026 \\
7 tags, groomer turn, semantic & 0.768 $\pm$ 0.019 & 0.748 \\
4 phases, exchange (G+V), lexical & 0.925 $\pm$ 0.009 & 0.919 $\pm$ 0.010 \\
4 phases, groomer turn only, lexical & 0.880 $\pm$ 0.016 & 0.869 $\pm$ 0.018 \\
4 phases, groomer turn, semantic & 0.829 $\pm$ 0.007 & 0.819 \\
4 phases, exchange, BERT fine-tuned & 0.918 $\pm$ 0.014 & 0.907 $\pm$ 0.013 \\
8 tags, v1.0, random folds, lexical & 0.660 $\pm$ 0.007 & 0.671 $\pm$ 0.010 \\
8 tags, v1.0, duplicates grouped, lexical & 0.735 $\pm$ 0.011 & 0.732 $\pm$ 0.012 \\
Majority class (7 tags / 4 phases) & 0.168 / 0.410 & --- \\
\bottomrule
\end{tabular}
\end{table}

\subsection{External validation against real ChatCoder data}
\label{sec:external}
Real lines were lowercased, and the ChatCoder netspeak translation was partially reverted (e.g. the 'question' suffixes were restored to question marks and 'laugh out loud' to 'lol'). Synthetic groomer messages were lowercased and stripped of emojis. The offender/groomer text was compared; the PJ victim side comprised adult decoys. For the semantic experiments, the real silver set was capped at 800 lines per tag (5,230 lines), while the gold set was sampled to comprise 4,000 posts.

\begin{figure*}[!t]
\centering
\includegraphics[width=.95\textwidth]{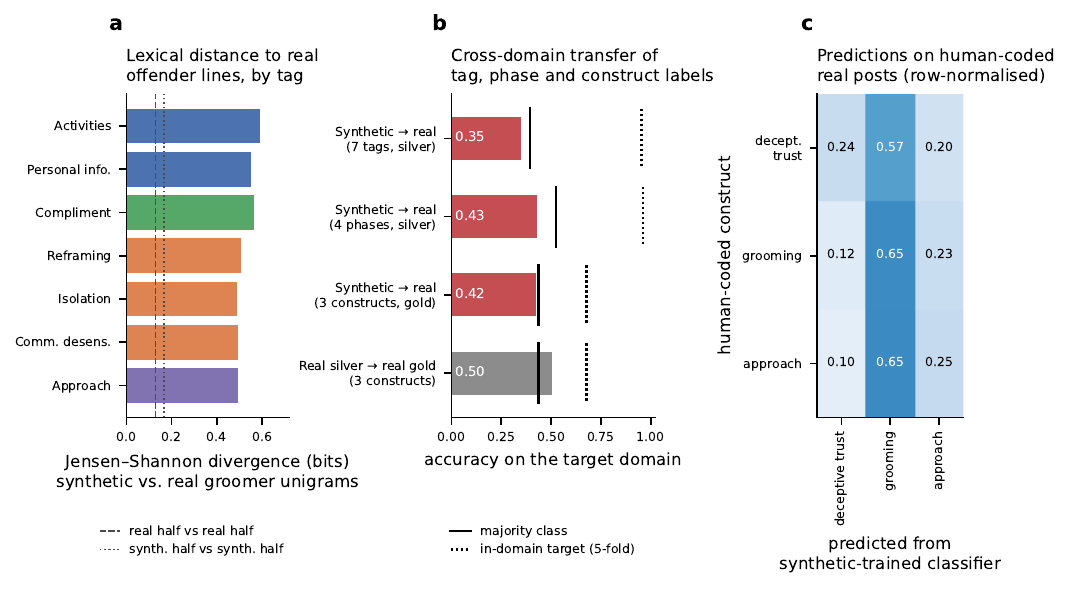}
\caption{External validation against real ChatCoder data (lexical features): (a) The Jensen-Shannon divergence between the unigram distributions of real and synthetic offender data per tag. The dashed and dotted lines show the divergence between two random halves of the real and synthetic data, respectively. (b) Accuracy of classifiers trained in one domain and tested in the other (bars), compared with the target majority class (solid tick) and the target's own in-domain 5-fold accuracy (dotted tick). (c) A row-normalised confusion matrix of the synthetic-trained classifier on human-coded constructs (n = 10,876 posts).}
\label{fig:external}
\end{figure*}

\textbf{Distributional distance.} The Jensen–Shannon divergence between the unigram distributions of the synthetic and real offender lines per tag ranges from 0.49 bits (Isolation and Communicative Desensitisation) to 0.59 bits (Activities). The divergence between two random halves of the real data is 0.13, and between two halves of the synthetic data it is 0.17 (Fig.~\ref{fig:external}a). When pooled over tags, the divergence is 0.45. Between 26\% (Activities) and 85\% (Communicative Desensitisation) of a tag's synthetic vocabulary occurs in the real lines of the same tag, and the mean lengths are comparable (7.8–9.9 vs. 6.4–9.3 tokens).

\textbf{Cross-domain transfer.} Table~\ref{tab:transfer} and Fig.~\ref{fig:external}b report classifiers trained on one domain and tested on the other. With lexical features, a classifier trained on synthetic groomer messages achieves an accuracy of 0.348 on the seven real tags and 0.427 on the four phases (real majority: 0.394 and 0.524; in-domain real: 0.953 and 0.961, with high figures reflecting the lexical origin of the silver labels) and 0.422 on the three gold constructs (majority: 0.437; label-permutation floor: 0.278; in-domain gold: 0.677 ± 0.019). The highest per-class recall on real lines is for Isolation (0.59) and Activities (0.66). Using semantic features, the same transfers achieve 0.388 accuracy for tags (majority 0.153 on the tag-capped sample), 0.448 accuracy for phases (majority 0.459), and 0.494 accuracy for constructs (majority 0.432; in-domain gold 0.734 ± 0.017). In the reverse direction, real silver lines recover the synthetic tags with 26.8\% (lexical) and 41.5\% (semantic) accuracy. Real silver labels transfer to gold labels with 50.3\% (lexical) and 56.7\% (semantic) accuracy, placing construct-level coding of real chat among the more challenging tasks with this material. The confusion matrix on the gold constructs (Fig.~\ref{fig:external}c) shows 57-65\% of real posts of every construct assigned to grooming, the construct corresponding to the largest synthetic phase.

\begin{table}[!t]
\caption{Cross-domain transfer between the synthetic corpus and real ChatCoder offender lines (classifier trained on the source, tested on the target). Silver = eight-tag dictionary coding of 288 transcripts (31,640 single-tag lines; 5,230 in the semantic setting); gold = human three-construct coding of 55 transcripts (10,876 posts; 4,000 in the semantic setting). The last column is the target's own 5-fold accuracy with transcript-grouped folds.}
\label{tab:transfer}
\centering\footnotesize
\begin{tabular}{@{}lcccc@{}}
\toprule
\textbf{Direction / label set} & \textbf{Acc.} & \textbf{F1} & \textbf{Maj.} & \textbf{In-dom.} \\
\midrule
\multicolumn{5}{@{}l}{\textit{Lexical (TF--IDF + LR)}} \\
Synthetic $\to$ real, 7 tags (silver) & 0.348 & 0.291 & 0.394 & 0.953 \\
Synthetic $\to$ real, 4 phases (silver) & 0.427 & 0.318 & 0.524 & 0.961 \\
Synthetic $\to$ real, 3 constructs (gold) & 0.422 & 0.371 & 0.437 & 0.677 \\
Real (silver) $\to$ synthetic, 7 tags & 0.268 & 0.222 & 0.168 & 0.788 \\
Real (silver) $\to$ synthetic, 4 phases & 0.422 & 0.309 & 0.410 & 0.880 \\
Real silver $\to$ real gold, 3 constructs & 0.503 & 0.418 & 0.437 & 0.677 \\
\midrule
\multicolumn{5}{@{}l}{\textit{Semantic (frozen all-roberta-large-v1 + LR)}} \\
Synthetic $\to$ real, 7 tags (silver) & 0.388 & 0.375 & 0.153 & 0.869 \\
Synthetic $\to$ real, 4 phases (silver) & 0.448 & 0.367 & 0.459 & --- \\
Synthetic $\to$ real, 3 constructs (gold) & 0.494 & 0.456 & 0.432 & 0.734 \\
Real (silver) $\to$ synthetic, 7 tags & 0.415 & 0.396 & 0.168 & 0.768 \\
Real silver $\to$ real gold, 3 constructs & 0.567 & 0.531 & 0.432 & 0.734 \\
\bottomrule
\end{tabular}
\end{table}

\textbf{Real-versus-synthetic discriminator.} A lexical classifier trained to distinguish synthetic grooming messages from a random sample of real offender messages of the same length achieves an area under the curve (AUC) of 0.962 and an accuracy of 0.906 (5-fold). On sentence embeddings, the AUC is 0.944. The words that are most indicative of synthetic content are contracted forms and the generator's register (e.g. 'you're', 'ur', 'hey', 'favorite', 'always', 'wanna', 'such', 'anyone', 'private', 'secret', 'game'), while the words that are most indicative of real content are endearments, body words and logistics (e.g. 'baby', 'sweetie', 'hun', 'bed', 'face', 'age', 'number', 'picture', 'camera'). The two registers can be separated on lexical and semantic grounds, independently of emoji and spelling. The synthetic corpus is polite, question-driven, and gaming-related, while the 2004–2008 PJ material is driven by endearments and logistics.

\section{Stage 5: Documentation, Scope of Use and Protocol Refinements}
\label{sec:discussion}
\subsection{Scope of use}
The measurements of Section~\ref{sec:validation} define the corpus. Its tags and phases are consistently expressed in the text (recoverability: 0.85/0.93). It contains no data from minors and no offender lines from the flagged subset of the seed material. Its register is characteristic of the generator and separable from the 2004–2008 PJ offender chat register (discriminator AUC: 0.94-0.96). Labels learned on it transfer to real offender lines with an accuracy of 0.35-0.49, which is above the majority class on the gold constructs with semantic features. The corpus therefore supports controlled experiments in which the phase is the manipulated variable, as well as pipeline, prompt and annotation-tool development. It can also be used to augment real data alongside the transfer gap being measured on the target, and for educational and awareness material where the clarity of the manipulative move is the relevant property. The datasheet outlines this scope and the transfer figures provide a reference point for any new use.

\subsection{Protocol refinements}
Applying the generation protocol \cite{franzoni2025responsible} to this corpus yields five refinements for future synthetic corpora in sensitive domains. 1) \emph{Integrity first}: byte-wise comparison of output files and normalised de-duplication precede every other measurement, and per-label counts are reported exactly.

2) \emph{Measured, flagged seed reproduction}: an instruction to paraphrase is complemented by a measurement of reproduction against the retrieval base and by row-level flags; reproduction concentrates in the tags where the generator's own knowledge is thinnest.

3) \emph{Recoverability as a consistency check}: in-domain accuracy documents label consistency, and a lexical model provides the reference against which transformer results are read.

4) \emph{Real reference at its own granularity}: each available real resource is used with its own label set and label status stated (silver dictionary coding, gold human coding), under both lexical and semantic features. 

5) \emph{Run logging}: decoding parameters, seeds and per-model refusal counts are recorded per run; the released code implements this for future generations.

\subsection{Limitations}
The real reference consists of decoy conversations from 2004–2008, which were coded using either a keyword dictionary (silver) or a numeric codebook that was not distributed with the data (gold). The construct mapping was inferred from the co-occurrence of dictionary tags with each code, as well as from the theory. The synthetic corpus is English-only, text-only and static. It was produced using one 8B checkpoint at 4-bit quantisation. A larger generator or one adapted to the domain would change the diversity and transfer figures. The protocol therefore prescribes measuring them again. The emotion field is silver annotation. Seed-reproduction analysis covers the 337 curated seed lines, but cannot reproduce other PJ material memorised during pre-training.

\subsection{Toward dynamic, role-aware successors}
This study has three key findings:

1) Emotion-trajectory features across the turns of a conversation, exploiting the emotion field released with the corpus.

2) A two-agent simulator in which domain-adapted models play groomer and victim to generate richer synthetic dialogues, as well as serving as an awareness-training tool for parents, educators, and minors.

3) Domain-specific fine-tuning to replace prompt-level control of general-purpose checkpoints. The present corpus and its validation figures serve as a reference point for this agenda.

\section{Reproducibility and Artifact Release}
\label{sec:repro}
We release, under the vetted-access licence: the labelled corpus (v1.1 and v1.2) with datasheet; the structure and formatting specification of the five phase documents; redacted system and question prompt templates; the generation protocol (strategy, batch size, model identifiers and quantisations); the real-reference construction, audit and validation code with the result tables; and the complete classifier configurations, which suffice to reproduce Tables~\ref{tab:recov}--\ref{tab:transfer} on the released data. Generation ran on the UniNuvola node described in Section~\ref{sec:method} with Python 3.10, PyTorch 2.2.1, Hugging Face Transformers 4.38.2, CUDA 12.1, Ollama 0.4.6 and LangChain 0.2.2. Random seeds for exact regeneration are not released, as part of the dual-use mitigation that governs the licence; seed documents, prompt components addressing refusal, and environment files are likewise withheld. The ChatCoder resources used as real references were distributed by their authors \cite{chatcoder}.

\section{Ethical Considerations and Responsible Disclosure}
\label{sec:ethics}
This research concerns sexual communication with minors and is classified as dual-use. The generation process was localised and the only available material is the publicly distributed Perverted Justice archive. In this archive, the victims are adult volunteers. The seed documents contain excerpts of real offender messages and remain with the authors. The prompt components that address refusal are described functionally. The corpus is released under a restricted, vetted-access licence to verified researchers and safeguarding organisations, along with the datasheet, audit code and v1.2 flags, under a data-use agreement. Samples of the most explicit tags are included and flagged because they are required for validation purposes. This work was conducted for preventive, educational and law enforcement support purposes under institutional oversight.

\section*{Data Availability}
A partial version of the dataset will be deposited in IEEE DataPort (\url{https://ieee-dataport.org/documents/lure-luring-theory-utterances-research-grooming-exchanges}) and made available in a limited version upon acceptance. The full dataset will be made available to verified researchers and safeguarding organisations upon vetted access, under a data-use agreement. During the review process, the dataset and datasheet can be provided to the editorial office or reviewers via an appropriate channel. Seed materials, random seeds for exact regeneration and components for handling refusals are not released. Requests can be directed to the corresponding author.

\section*{Acknowledgment}
This paper builds on the M.Sc.\ thesis of L.~Guercio (University of Perugia) \cite{guercio2024thesis}. The authors thank the stakeholders in child-safety and law enforcement who advised on the phase taxonomy and ethical governance. This work did not receive project funding; it is supported by the Emotions and Multidisciplinary-Oriented Research on Ethical and Responsible AI (EMORE) Lab, University of Perugia, Italy. The OA publication is funded by the University of Perugia. The authors declare the use of Large Language Models as a research methodology for generating the synthetic dataset under human supervision, as detailed in Sections~\ref{sec:method}-\ref{sec:hitl}; all content has been verified by the authors, who assume full responsibility for this work. In line with current editorial policies on the use of artificial intelligence in article preparation, the authors disclose that an AI assistant (Claude Science) was used to support coding for data analysis and for language enhancement (DeepL Write) under the authors' direction; all analyses, results and text were checked by the authors. The authors declare no conflict of interest.

\section*{Author Contributions}
Authors are listed in alphabetical order. Conceptualization: V.F.; Methodology: V.F., A.M., L.G.; Software: L.G.; Validation: V.F., L.G., E.F., M.P.; Formal analysis: V.F., L.G.; Investigation: L.G.; Data curation: L.G., M.P.; Visualization: L.G., V.F.; Writing-original draft: V.F., L.G., M.P.; Writing-review and editing: V.F., M.P., E.F.; Supervision: V.F.

All authors have read and agreed to the published version of the manuscript.

\bibliographystyle{IEEEtran}
\bibliography{biblio_tifs}

@ARTICLE{10948513,
  author={Ghadermazi, Jalal and Hore, Soumyadeep and Shah, Ankit and Bastian, Nathaniel D.},
  journal={IEEE Transactions on Information Forensics and Security}, 
  title={GTAE-IDS: Graph Transformer-Based Autoencoder Framework for Real-Time Network Intrusion Detection}, 
  year={2025},
  volume={20},
  number={},
  pages={4026-4041},
  doi={10.1109/TIFS.2025.3557741}}

@ARTICLE{10964328,
  author={Shen, Meng and Wu, Jinhe and Ye, Ke and Xu, Ke and Xiong, Gang and Zhu, Liehuang},
  journal={IEEE Transactions on Information Forensics and Security}, 
  title={Robust Detection of Malicious Encrypted Traffic via Contrastive Learning}, 
  year={2025},
  volume={20},
  number={},
  pages={4228-4242},
  doi={10.1109/TIFS.2025.3560560}}

@ARTICLE{11003860,
  author={Chan, Alyssa and Hallett, Joseph and Edwards, Matthew},
  journal={IEEE Security \& Privacy}, 
  title={Why Is Johnny Getting Groomed in Roblox?}, 
  year={2025},
  volume={23},
  number={3},
  pages={20-30},
  doi={10.1109/MSEC.2025.3564542}}

@ARTICLE{11021376,
  author={Mujtaba, Ghulam and Ali Khowaja, Sunder and Dev, Kapal},
  journal={IEEE Internet of Things Journal}, 
  title={EdgeAIGuard: Agentic LLMs for Minor Protection in Digital Spaces}, 
  year={2025},
  volume={12},
  number={17},
  pages={34992-35000},
  doi={10.1109/JIOT.2025.3574961}}

@ARTICLE{8509614,
  author={Galbally, Javier and Haraksim, Rudolf and Beslay, Laurent},
  journal={IEEE Transactions on Information Forensics and Security}, 
  title={A Study of Age and Ageing in Fingerprint Biometrics}, 
  year={2019},
  volume={14},
  number={5},
  pages={1351-1365},
  doi={10.1109/TIFS.2018.2878160}}

@ARTICLE{9541357,
  author={Jevremovic, Aleksandar and Veinovic, Mladen and Cabarkapa, Milan and Krstic, Marko and Chorbev, Ivan and Dimitrovski, Ivica and Garcia, Nuno and Pombo, Nuno and Stojmenovic, Milos},
  journal={IEEE Access}, 
  title={Keeping Children Safe Online With Limited Resources: Analyzing What is Seen and Heard}, 
  year={2021},
  volume={9},
  number={},
  pages={132723-132732},
  doi={10.1109/ACCESS.2021.3114389}}

@misc{Chase2022LangChain,
  author = {Harrison Chase},
  title = {LangChain},
  year = {2022},
  howpublished = {\url{https://github.com/langchain-ai/langchain}},
  note = {Accessed: 2025-07-09}
}

@ARTICLE{Cheong2015220,
	author = {Cheong, Yun-Gyung and Jensen, Alaina K. and Gudnadottir, Elin Rut and Bae, Byung-Chull and Togelius, Julian},
	title = {Detecting Predatory Behavior in Game Chats},
	year = {2015},
	journal = {IEEE Transactions on Computational Intelligence and AI in Games},
	volume = {7},
	number = {3},
	pages = {220 – 232},
	doi = {10.1109/TCIAIG.2015.2424932},
	type = {Article},
	publication_stage = {Final},
	source = {Scopus},
	note = {Cited by: 24}
}

@misc{Huihui2025DeepSeekAbliterated,
  author       = {huihui-ai},
  title        = {DeepSeek-R1-Distill-Llama-8B-abliterated},
  year         = {2025},
  howpublished = {\url{https://huggingface.co/huihui-ai/DeepSeek-R1-Distill-Llama-8B-abliterated}},
  note         = {Hugging Face model card. Accessed: 2025-07-09}
}

@misc{Huihui2025Llama32Abliterated,
  author       = {huihui-ai},
  title        = {Llama-3.2-3B-Instruct-abliterated},
  year         = {2025},
  howpublished = {\url{https://huggingface.co/huihui-ai/Llama-3.2-3B-Instruct-abliterated}},
  note         = {Hugging Face model card. Accessed: 2025-07-09}
}

@misc{Labonne2024Llama31Abliterated,
  author       = {Maxime Labonne},
  title        = {Meta-Llama-3.1-8B-Instruct-abliterated},
  year         = {2024},
  howpublished = {\url{https://huggingface.co/mlabonne/Meta-Llama-3.1-8B-Instruct-abliterated}},
  note         = {Hugging Face model card. Accessed: 2025-07-09}
}

@ARTICLE{Murnion2018197,
	author = {Murnion, Shane and Buchanan, William J. and Smales, Adrian and Russell, Gordon},
	title = {Machine learning and semantic analysis of in-game chat for cyberbullying},
	year = {2018},
	journal = {Computers and Security},
	volume = {76},
	pages = {197 – 213},
	doi = {10.1016/j.cose.2018.02.016},
	type = {Article},
	publication_stage = {Final},
	source = {Scopus},
	note = {Cited by: 78; All Open Access, Green Open Access}
}

@misc{Ollama2024,
  author = {{Ollama}},
  title = {Ollama},
  year = {2024},
  howpublished = {\url{https://ollama.com}},
  note = {Accessed: 2025-07-09}
}

@CONFERENCE{Yoo20212225,
	author = {Yoo, Kang Min and Park, Dongju and Kang, Jaewook and Lee, Sang-Woo and Park, Woomyeong},
	title = {GPT3Mix: Leveraging Large-scale Language Models for Text Augmentation},
	year = {2021},
	journal = {Findings of the Association for Computational Linguistics, Findings of ACL: EMNLP 2021},
	pages = {2225 – 2239},
	doi = {10.18653/v1/2021.findings-emnlp.192},
	affiliations = {NAVER AI Lab, South Korea; NAVER Clova AI, South Korea},
	editor = {Moens M.-F. and Huang X. and Specia L. and Yih S.W.-T.},
	publisher = {Association for Computational Linguistics (ACL)},
	isbn = {978-195591710-0},
	language = {English},
	abbrev_source_title = {Find. Assoc. Comput. Linguist., Findings ACL: EMNLP},
	type = {Conference paper},
	publication_stage = {Final},
	source = {Scopus},
	note = {Cited by: 158; All Open Access, Gold Open Access},
    booktitle = {Findings of ACL: EMNLP 2021 November 7-11, 2021}
}

@article{affective1,
  author  = {Franzoni, Valentina and Biondi, Giulio and Milani, Alfredo},
  title   = {Emotional Sounds of Crowds: Spectrogram-Based Analysis Using Deep Learning},
  journal = {Multimedia Tools and Applications},
  volume  = {79},
  number  = {47-48},
  pages   = {36063--36075},
  year    = {2020},
  doi     = {10.1007/s11042-020-09428-x}
}

@article{affective2,
  author  = {Baia, Alina Elena and Biondi, Giulio and Franzoni, Valentina and Milani, Alfredo and Poggioni, Valentina},
  title   = {Lie to Me: Shield Your Emotions from Prying Software},
  journal = {Sensors},
  volume  = {22},
  number  = {3},
  pages   = {967},
  year    = {2022},
  doi     = {10.3390/s22030967}
}

@article{amici2026plumed,
  author  = {Amici, Marco and Polticchia, Mattia and Florindi, Emanuele and Franzoni, Valentina},
  title   = {A Synthetic {P}lutchik-Grounded Corpus for Fine-Grained Emotion Recognition in Text},
  journal = {Neural Computing and Applications},
  year    = {2026},
  note    = {Manuscript NCAA-D-26-04319, under review, can be put in Arxiv if acceptance doesn't arrive before this paper}
}

@inproceedings{arditi2024refusal,
  author    = {Arditi, Andy and Obeso, Oscar and Syed, Aaquib and Paleka, Daniel and Panickssery, Nina and Gurnee, Wes and Nanda, Neel},
  title     = {Refusal in Language Models Is Mediated by a Single Direction},
  booktitle = {Advances in Neural Information Processing Systems},
  volume    = {37},
  pages     = {136037--136083},
  year      = {2024},
  doi       = {10.52202/079017-4322}
}

@article{bender2018data,
  author  = {Bender, Emily M. and Friedman, Batya},
  title   = {Data Statements for Natural Language Processing: Toward Mitigating System Bias and Enabling Better Science},
  journal = {Transactions of the Association for Computational Linguistics},
  volume  = {6}, pages = {587--604}, year = {2018},
  doi     = {10.1162/tacl_a_00041}
}

@inproceedings{chatcoder,
  author    = {Kontostathis, April and Edwards, Lynne and Leatherman, Amanda},
  title     = {{ChatCoder}: Toward the Tracking and Categorization of Internet Predators},
  booktitle = {Proceedings of the Text Mining Workshop, Ninth {SIAM} International Conference on Data Mining ({SDM} 2009)},
  year      = {2009}
}

@inproceedings{chehbouni2024federated,
  author    = {Chehbouni, Khaoula and De Cock, Martine and Caporossi, Gilles and Taik, Afaf and Rabbany, Reihaneh and Farnadi, Golnoosh},
  title     = {Enhancing Privacy in the Early Detection of Sexual Predators Through Federated Learning and Differential Privacy},
  booktitle = {Proc. AAAI Conference on Artificial Intelligence},
  volume    = {39},
  number    = {27},
  pages     = {27887--27895},
  year      = {2025},
  doi       = {10.1609/aaai.v39i27.35005}
}

@book{cialdini2009,
  author    = {Cialdini, Robert B.},
  title     = {Influence: The Psychology of Persuasion},
  publisher = {HarperCollins},
  year      = {2009},
  isbn      = {9780061899874}
}

@inproceedings{dataaug,
  author    = {Long, Lin and Wang, Rui and Xiao, Ruixuan and Zhao, Junbo and Ding, Xiao and Chen, Gang and Wang, Haobo},
  title     = {On {LLMs}-Driven Synthetic Data Generation, Curation, and Evaluation: A Survey},
  booktitle = {Findings of the Association for Computational Linguistics: ACL 2024},
  address   = {Bangkok, Thailand},
  publisher = {Association for Computational Linguistics},
  pages     = {11065--11082},
  year      = {2024},
  doi       = {10.18653/v1/2024.findings-acl.658}
}

@article{ekman1992,
  author  = {Ekman, Paul},
  title   = {An Argument for Basic Emotions},
  journal = {Cognition and Emotion},
  volume  = {6},
  number  = {3-4},
  pages   = {169--200},
  year    = {1992},
  doi     = {10.1080/02699939208411068}
}

@book{finkelhor1984,
  author    = {Finkelhor, David},
  title     = {Child Sexual Abuse: New Theory and Research},
  publisher = {Free Press},
  address   = {New York, NY, USA},
  year      = {1984}
}

@article{florindi2024,
  author  = {Florindi, Emanuele},
  title   = {Pedofilia e videogiochi: storia ed evoluzione di un fenomeno ({P}edophilia and Video Games: Analysis of a Growing Phenomenon)},
  journal = {SSRN},
  year    = {2024},
  doi     = {10.2139/ssrn.4736912}
}

@inproceedings{franzoni2025responsible,
  author    = {Franzoni, Valentina},
  title     = {Towards Responsible Generation of Synthetic Datasets with Large Language Models},
  booktitle = {Proceedings of the 2025 IEEE/WIC International Conference on Web Intelligence and Intelligent Agent Technology (WI-IAT)},
  publisher = {IEEE}, year = {2025},
  doi       = {10.1109/WI-IAT67162.2025.00123},
  note      = {VERIFY co-authors and page numbers against the IEEE Xplore record (document 11676877)}
}

@misc{gdpr,
  author       = {{European Union}},
  title        = {Regulation ({EU}) 2016/679 (General Data Protection Regulation); Directive ({EU}) 2016/680},
  howpublished = {Official Journal of the European Union, L~119},
  year         = {2016},
  url          = {https://eur-lex.europa.eu/legal-content/EN/TXT/?uri=CELEX:32016R0679}
}

@article{gebru2021datasheets,
  author  = {Gebru, Timnit and Morgenstern, Jamie and Vecchione, Briana and Vaughan, Jennifer Wortman and Wallach, Hanna and Daum{\'e} III, Hal and Crawford, Kate},
  title   = {Datasheets for Datasets},
  journal = {Communications of the ACM},
  volume  = {64}, number = {12}, pages = {86--92}, year = {2021},
  doi     = {10.1145/3458723}
}

@article{groomingsurvey,
  author  = {Borj, Parisa Rezaee and Raja, Kiran and Bours, Patrick},
  title   = {Online grooming detection: A comprehensive survey of child exploitation in chat logs},
  journal = {Knowledge-Based Systems},
  volume  = {259},
  pages   = {110039},
  year    = {2023},
  doi     = {10.1016/j.knosys.2022.110039}
}

@techreport{iwf,
  author      = {{Internet Watch Foundation}},
  title       = {Annual Report 2023},
  institution = {Internet Watch Foundation},
  address     = {Cambridge, U.K.},
  year        = {2023},
  url         = {https://www.iwf.org.uk}
}

@inproceedings{li2023synthetic,
  author    = {Li, Zhuoyan and Zhu, Hangxiao and Lu, Zhuoran and Yin, Ming},
  title     = {Synthetic Data Generation with Large Language Models for Text Classification: Potential and Limitations},
  booktitle = {Proceedings of the 2023 Conference on Empirical Methods in Natural Language Processing (EMNLP)},
  year      = {2023}, note = {VERIFY pages/DOI; arXiv:2310.07849}
}

@article{lorenzodus2016understanding,
  author  = {Lorenzo-Dus, Nuria and Izura, Cristina and P{\'e}rez-Tattam, Roc{\'i}o},
  title   = {Understanding Grooming Discourse in Computer-Mediated Environments},
  journal = {Discourse, Context \& Media},
  volume  = {12},
  pages   = {40--50},
  year    = {2016},
  doi     = {10.1016/j.dcm.2016.02.004}
}

@misc{ncmec,
  author       = {{National Center for Missing \& Exploited Children}},
  title        = {CyberTipline Reports and Data},
  howpublished = {\url{https://www.missingkids.org/cybertiplinedata}},
  address      = {Alexandria, VA, USA},
  year         = {2023}
}

@misc{neweulaw26,
  author = {{European Parliament and Council}},
  title = {Regulation ({EU}) 2021/1232 on a temporary derogation for the purpose of combating online child sexual abuse, and the proposed Regulation laying down rules to prevent and combat child sexual abuse ({COM}(2022) 209)},
  howpublished = {\url{https://eur-lex.europa.eu/legal-content/EN/TXT/?uri=CELEX:32021R1232}},
  year = {2021},
  note = {VERIFY: cite the 2026 legislative act/position actually referred to}
}

@article{nrclexicon,
  author  = {Mohammad, Saif M. and Turney, Peter D.},
  title   = {Crowdsourcing a Word--Emotion Association Lexicon},
  journal = {Computational Intelligence},
  volume  = {29},
  number  = {3},
  pages   = {436--465},
  year    = {2013},
  doi     = {10.1111/j.1467-8640.2012.00460.x}
}

@article{olson2007luring,
  author  = {Olson, Loreen N. and Daggs, Joy L. and Ellevold, Barbara L. and Rogers, Teddy K. K.},
  title   = {Entrapping the Innocent: Toward a Theory of Child Sexual Predators' Luring Communication},
  journal = {Communication Theory},
  volume  = {17}, number = {3}, pages = {231--251}, year = {2007},
  doi     = {10.1111/j.1468-2885.2007.00294.x}
}

@inproceedings{pan12,
  author    = {Inches, Giacomo and Crestani, Fabio},
  title     = {Overview of the International Sexual Predator Identification Competition at {PAN}-2012},
  booktitle = {CLEF 2012 Evaluation Labs and Workshop -- Working Notes Papers},
  series    = {CEUR Workshop Proceedings},
  volume    = {1178},
  year      = {2012}
}

@misc{pj,
  author       = {{Perverted Justice Foundation}},
  title        = {Perverted-{J}ustice Chat-Log Archive},
  howpublished = {Archived chat logs},
  year         = {2007}
}

@misc{plutchickTextCorpus,
  author = {Amici, Marco and Polticchia, Mattia and Florindi, Emanuele and Franzoni, Valentina},
  title={A Synthetic Plutchik-Grounded Corpus for Fine-Grained Emotion Recognition in Text},
  howpublished = {IEEE DataPort},
  year={2026},
  url={https://ieee-dataport.org/documents/synthetic-plutchik-grounded-corpus-fine-grained-emotion-recognition-text}
}

@article{plutchik2001,
  author  = {Plutchik, Robert},
  title   = {The Nature of Emotions: Human Emotions Have Deep Evolutionary Roots, a Fact That May Explain Their Complexity and Provide Tools for Clinical Practice},
  journal = {American Scientist},
  volume  = {89},
  number  = {4},
  pages   = {344--350},
  year    = {2001}
}

@techreport{polposta2024,
  author      = {{Polizia Postale e delle Comunicazioni}},
  title       = {Resoconto annuale delle attivit{\`a} contro la pedofilia online e i reati contro i minori},
  institution = {Polizia di Stato},
  address     = {Rome, Italy},
  year        = {2024},
  url         = {https://www.poliziadistato.it}
}

@inproceedings{prosser2024helpful,
  author    = {Prosser, Ellie and Edwards, Matthew},
  title     = {Helpful or Harmful? {E}xploring the Efficacy of Large Language Models for Online Grooming Prevention},
  booktitle = {Proc. European Interdisciplinary Cybersecurity Conference (EICC 2024)},
  publisher = {ACM},
  address   = {New York, NY, USA},
  pages     = {1--10},
  year      = {2024},
  doi       = {10.1145/3655693.3655694}
}

@inproceedings{rag,
 author = {Lewis, Patrick and Perez, Ethan and Piktus, Aleksandra and Petroni, Fabio and Karpukhin, Vladimir and Goyal, Naman and K\"{u}ttler, Heinrich and Lewis, Mike and Yih, Wen-tau and Rockt\"{a}schel, Tim and Riedel, Sebastian and Kiela, Douwe},
 booktitle = {Advances in Neural Information Processing Systems},
 editor = {H. Larochelle and M. Ranzato and R. Hadsell and M.F. Balcan and H. Lin},
 pages = {9459--9474},
 publisher = {Curran Associates, Inc.},
 title = {Retrieval-Augmented Generation for Knowledge-Intensive NLP Tasks},
 volume = {33},
 year = {2020}
}

@techreport{ran2021,
  author      = {{Radicalisation Awareness Network (RAN C\&N)}},
  title       = {Digital Grooming Tactics on Video Gaming \& Video Gaming Adjacent Platforms: Threats and Opportunities},
  institution = {European Commission, DG Migration and Home Affairs},
  year        = {2021},
  url         = {https://home-affairs.ec.europa.eu/system/files/2021-05/ran_c-n_conclusion_paper_grooming_through_gaming_15-16032021_en.pdf}
}

@article{shumailov2024collapse,
  author  = {Shumailov, Ilia and Shumaylov, Zakhar and Zhao, Yiren and Papernot, Nicolas and Anderson, Ross and Gal, Yarin},
  title   = {{AI} Models Collapse When Trained on Recursively Generated Data},
  journal = {Nature}, volume = {631}, pages = {755--759}, year = {2024},
  doi     = {10.1038/s41586-024-07566-y}
}

@article{tyad020,
    author = {Abelson, Harold and Anderson, Ross and Bellovin, Steven M and Benaloh, Josh and Blaze, Matt and Callas, Jon and Diffie, Whitfield and Landau, Susan and Neumann, Peter G and Rivest, Ronald L and Schiller, Jeffrey I and Schneier, Bruce and Teague, Vanessa and Troncoso, Carmela},
    title = {Bugs in our pockets: the risks of client-side scanning},
    journal = {Journal of Cybersecurity},
    volume = {10},
    number = {1},
    pages = {tyad020},
    year = {2024},
    month = {01},
    issn = {2057-2085},
    doi = {10.1093/cybsec/tyad020},
 }

@inproceedings{vogt2021early,
  author    = {Vogt, Matthias and Leser, Ulf and Akbik, Alan},
  title     = {Early Detection of Sexual Predators in Chats},
  booktitle = {Proc. 59th Annual Meeting of the Association for Computational Linguistics and the 11th International Joint Conference on Natural Language Processing (Volume 1: Long Papers)},
  publisher = {Association for Computational Linguistics},
  pages     = {4985--4999},
  year      = {2021},
  doi       = {10.18653/v1/2021.acl-long.386}
}

@article{ward2001,
  author  = {Ward, Tony and Hudson, Stephen M.},
  title   = {Finkelhor's Precondition Model of Child Sexual Abuse: A Critique},
  journal = {Psychology, Crime \& Law},
  volume  = {7},
  number  = {4},
  pages   = {291--307},
  year    = {2001},
  doi     = {10.1080/10683160108401799}
}

@article{whittle2013,
  author  = {Whittle, Helen and Hamilton-Giachritsis, Catherine and Beech, Anthony and Collings, Guy},
  title   = {A Review of Young People's Vulnerabilities to Online Grooming},
  journal = {Aggression and Violent Behavior},
  volume  = {18},
  number  = {1},
  pages   = {135--146},
  year    = {2013},
  doi     = {10.1016/j.avb.2012.11.008}
}

@article{winters2017sexual,
  author  = {Winters, Georgia M. and Kaylor, Leah E. and Jeglic, Elizabeth L.},
  title   = {Sexual Offenders Contacting Children Online: An Examination of Transcripts of Sexual Grooming},
  journal = {Journal of Sexual Aggression},
  volume  = {23},
  number  = {1},
  pages   = {62--76},
  year    = {2017},
  doi     = {10.1080/13552600.2016.1271146}
}

@article{wolak2004,
  author  = {Wolak, Janis and Finkelhor, David and Mitchell, Kimberly},
  title   = {Internet-Initiated Sex Crimes Against Minors: Implications for Prevention Based on Findings from a National Study},
  journal = {Journal of Adolescent Health},
  volume  = {35},
  number  = {5},
  pages   = {424.e11--424.e20},
  year    = {2004}
}

@mastersthesis{guercio2024thesis,
  author  = {Guercio, Ludovico and Polticchia, Mattia and Florindi, Emanuele and Franzoni, Valentina},
  title   = {Towards the Prevention of Online Grooming: Arousal Classification in the Early Phases of Online Luring},
  school  = {Universit{\`a} degli Studi di Perugia, Dipartimento di Matematica e Informatica},
  address = {Perugia, Italy},
  year    = {2023/2024},
  type    = {M.Sc. thesis}
}

\begin{IEEEbiographynophoto}{Emanuele Florindi}
is a PhD candidate in Learning Sciences and Digital Technologies at the University of Modena and Reggio Emilia, Italy, previously adjunct professor in law and emerging technologies at the Department of Mathematics and Computer Science, University of Perugia, where he serves as Data Protection Officer. His research spans cybercrime, artificial intelligence and military professional training. He received his degrees in law and in psychology from the University of Perugia, is a member of scientific and ethical committees in international projects, and a member of the IEEE.
\end{IEEEbiographynophoto}

\begin{IEEEbiographynophoto}{Valentina Franzoni}
is Associate Professor at the Department of Mathematics and Computer Science, University of Perugia, Italy, where she serves as Rector's subdelegate for Artificial Intelligence, with responsibility for affective computing, AI ethics and social impact, and leads the Emotions and Multidisciplinary-Oriented Research on Ethical and Responsible AI (EMORE) Lab. Her research interests include complex networks, artificial intelligence and affective computing. She received her PhD in Engineering for Computer Science from Sapienza University of Rome (DIAG). She is a Senior Member of the IEEE and a member of AAAC, ACM and the CINI Smart Cities \& Communities National Laboratory.
\end{IEEEbiographynophoto}

\begin{IEEEbiographynophoto}{Ludovico Guercio}
is a postgraduate of the Department of Mathematics and Computer Science, University of Perugia, Italy, where he received his B.Sc.\ in Computer Science and M.Sc.\ in Computer Science--Cybersecurity. His research focuses on artificial intelligence and cybersecurity, in particular AI-based prevention of online child grooming.
\end{IEEEbiographynophoto}

\begin{IEEEbiographynophoto}{Alfredo Milani}
is Associate Professor at Link Campus University, Rome, Italy, where he directs the AI Multicognitive Research Laboratory (AIMLab), affiliated with the CAIRNE European network of AI research. His research interests include evolutionary algorithms, agent planning and generative AI. He received his degree in Information Science from the University of Pisa, Italy, and is a Senior Member of the IEEE.
\end{IEEEbiographynophoto}

\begin{IEEEbiographynophoto}{Mattia Polticchia}
is a PhD candidate in Learning Sciences and Digital Technologies at the University of Modena and Reggio Emilia, Italy. His research interests span game science and serious games, virtual reality and affective computing. He received his B.Sc.\ in Computer Science and M.Sc.\ in Computer Science--Artificial Intelligence from the University of Perugia, Italy, summa cum laude, and is a member of the IEEE.
\end{IEEEbiographynophoto}

\end{document}